# How Age Influences the Interpretation of Emotional Body Language in Humanoid Robots

ILARIA CONSOLI, CLAUDIO MATTUTINO, and CRISTINA GENA, Computer Science Dept., University of Turin, Italy

BERARDINA DE CAROLIS and GIUSEPPE PALESTRA, University of Bari Aldo Moro, Italy

This paper presents an empirical study investigating how individuals across different age groups, children, young and older adults, interpret emotional body language expressed by the humanoid robot NAO. The aim is to offer insights into how users perceive and respond to emotional cues from robotic agents, through an empirical evaluation of the robot's effectiveness in conveying emotions to different groups of users. By analyzing data collected from elderly participants and comparing these findings with previously gathered data from young adults and children, the study highlights similarities and differences between the groups, with younger and older users more similar but different from young adults.

CCS Concepts: • Human-centered computing → User models; • Computing methodologies → Cognitive robotics. Additional Key

## 1 INTRODUCTION

Non-verbal communication, a crucial component of human interaction, includes gestures, facial expressions, posture, and other behaviors that convey meaning without speech [29]. There is a general consensus that body movements and postures provide important cues for identifying emotional states, particularly when facial and vocal signals are unavailable [1]. Emotional Body Language (EBL) is rapidly emerging as a significant area of research within cognitive and affective neuroscience. According to De Gelder [10], numerous valuable insights into human emotion and its neurobiological foundations have been derived from the study of facial expressions. Indeed certain emotions are more effectively conveyed through facial expressions, while others are better communicated through body movements or a combination of both. Gestures provide observable cues that can be instrumental in recognizing and interpreting a user's emotional state, especially in the absence of verbal or facial signals.

The neurobiological foundations of EBL remain relatively underexplored. De Gelder defines EBL as an *emotion expressed through the entire body, involving coordinated movements and often a meaningful action*. This definition urges researchers to move beyond the traditional focus on facial expressions and instead consider the perception of movement and action, dimensions that have typically been examined in isolation rather than in direct relation to EBL.

Emotions can be expressed through various types of body language, such as posture, movement, and proxemics [6, 28]. *Posture* is a static position that the body assumes at specific moments. It has been shown that people are particularly sensitive to what posture conveys. Recent scientific studies [17] suggest that there is a distinct part of the brain dedicated to recognizing biological input related to body posture and movement.

According to Beck et al. [2] *movement* is a good indicator for understanding a person's emotional state, regardless of the presence or absence of facial or vocal expressions. Body movements refer not only to the act of moving itself but also to the manner in which the movement is performed; for example, speed is a factor that may influence how a movement is interpreted.

Finally *proxemics* refers to the distance between individuals in a social interaction. Although it is not considered an emotional expression in itself, it is a useful component in completing the representation of an interaction, making it more realistic.

In this paper we translate the EBL perspective to the study and research in Human-Robot Interaction (HRI). Although a humanoid robot is, in fact, an artificial entity, it may still be expected to express itself through body language [2]. The human brain is naturally predisposed to interpret non-verbal signals, such as gestures and sounds, as meaningful. In general, all these *non-linguistic expressions* can be employed to integrate an additional non-verbal communication channel in social robots [5], thereby enhancing their expressiveness.

In this paper, we present the replication of an experiment initially conducted with university students and then with primary school children. We replicated the same experiment with elderly participants. The main research questions addressed by this study are the following: *Is NAO's emotional body language correctly understood?* (RQ1). *Do users across different age groups interpret emotional body language differently?* (RQ2). In the past, we had already observed differences between the children and the university students, and we were interested in understanding the reactions of older users.

The paper is organized as follows: Section 2 reviews the state of the art in EBL within Human-Robot Interaction; Section 3 outlines the theoretical background underpinning the study; Section 4 details the experimental design, procedure, and results; Section 5 presents a comparative analysis across age groups; and the final section discusses the conclusions and implications for future research.

## 2  STATE OF THE ART

In the context of HRI, robots should not only be able to recognize and classify EBL, but also use EBL to express their own emotions [19]. While some robots utilize facial expression to display simplified emotional expressions, others, such as Nao, rely primarily on body language, as they lack the capability to convey emotions through facial expressions.

In any case, social robots must be socially intelligent to engage in natural, bidirectional and emotional communication with humans. McColl et al. [20] investigated the design of EBL for a human-like social robots using various body postures and movements identified in human emotion research. Experimental results demonstrated that participants could accurately recognize the robot's emotional body language for sadness, joy, anger, surprise, and boredom.

Beck et al. [2] report a case study involving the NAO robot interacting with children. Their findings suggest that body posture and head position can effectively convey emotions in child–robot interaction. Specifically, they indicate that the expressiveness of negative emotions (e.g., anger and sadness) can be enhanced by lowering the head, while the expressiveness of positive emotions (e.g., happiness, excitement, and pride) can be enhanced by raising the head. In another experiment they consider Valence, Arousal and Stance of emotional key poses [3]. They find that the upward head position was consistently associated with higher arousal ratings compared to a neutral or downward position. However, evaluations of valence (i.e., the perceived positivity or negativity of the emotion) and stance (i.e., the perceived approachability or aversiveness of the robot) were influenced by the interaction between head position and the specific emotion being expressed. Finally in a third experiment [4], results show that EBL displayed by an agent or a human is interpreted in a similar way in terms of recognition. Overall, they conclude that these experiments confirm that body language is a suitable modality for robots to express emotions and suggest that an affective space for body expressions could enhance the emotional expressiveness of humanoid robots.

Marmpena et al. [18] investigate how humans perceive EBL performed by the humanoid robot Pepper, adopting a dimensional approach to emotion representation. In the study, 20 participants evaluated 36 animated expressions based solely on body movements, sometimes accompanied by LED or sound effects. The results revealed that arousal was generally easier to evaluate than valence, particularly in the absence of contextual or facial cues. Animations perceived as conveying high arousal tended to be more clearly identified, while lower-arousal animations were often rated ambiguously or shifted toward neutral valence.

Embgen et al. [13] present a pilot study aimed not at simulating emotions in a traditional sense, but at exploring whether users could interpret "abstracted" robot expressions (namely robot-specific affect communication) as specific emotional states. To this end, they introduced Daryl, a mobile robot with a mildly humanized design, and developed six motion sequences that integrated human-like, animal-like, and robot-specific social cues. The user study (N=29) revealed that, even in the absence of facial expressions and articulated limbs, participants' interpretations of Daryl's emotional states aligned with the intended abstract emotion displays. These findings suggest that such abstract, multimodal expressions may serve as a viable alternative to complex facial animation, and contribute to ongoing research on identifying robot-specific social cues.

McColl and Nejat [21] investigate the use of body movement and posture descriptors—drawn from human emotion research—for the human-like social robot Brian 2.0. They conduct experiments to assess whether non-expert individuals can recognize eight social emotions—sadness, fear, elated joy, surprise, anger, boredom, interest, and happiness—based solely on Brian 2.0's body language, and to compare the interpretation of these expressions when performed by the robot and a human actor. Their findings indicate that participants were able to accurately recognize emotions such as sadness, elated joy, anger, surprise, and boredom from the robot's body language. Notably, sadness was identified more successfully when expressed by the robot than by the human actor, and similar recognition rates were observed for elated joy, surprise, and interest across both agents. However, happiness was consistently the least recognized emotion for both, likely due to overlapping movement features with other emotions. Anger, fear, and boredom were more accurately recognized when expressed by the human actor. These results demonstrate that a life-sized human-like robot can effectively convey several social emotions through selected human-based body movements and postures.

## 3  BACKGROUND

A relevant part of the research on human-centered interaction with robot is focused on emotion [26, 27], both human emotion recognition and robot emotion expression through face and emotional body language. As outlined in Gaudi et al. [14]. emotion play a central role in social interaction and are involved in many other processes, including decision making.

In the broader field of affective computing [22], computational modeling of emotion has been addressed primarily by two approaches [25]: the *categorical approach* that categorizes emotions into mood classes and the *continuous dimensional approach* that usually regards emotions as numerical values over a few dimensions such as valence, arousal, and dominance. The categorical theory proposes the existence of basic, distinct and universal emotions, with clear boundaries separating emotional states. The dimensional theory proposes the existence of the fundamental dimensions that form an emotional space.

Understanding how clearly humans perceive a robot's EBL is essential for designing interactions that are as natural and fluent as possible. In the last years, we investigated this aspect by implementing gestures associated with specific emotions on the social robot NAO [8, 9]. We adopted a classical categorical framework, drawing on Ekman's model of basic emotions [12] as well as Plutchik's Wheel of Emotions [23], another widely recognized categorical approach.

The objective of the experiments was to determine whether emotions expressed by NAO through body language could be accurately recognized and classified by users. In the first experiment [9], following an initial session of gesture refinement based on user feedback, we involved a total of 20 university students, comprising 11 males and 9 females. In the second experiment [8] we involved 176 children participating to an educational robotic training [7, 16], comprising 86 males and 90 females, following the same procedures that will be described below.

In the following, we will describe the experimental approach and we will report the results of a new experiment conducted with elderly participants, then we will compare the results with the ones previously obtained with students and children.

## 4 THE EXPERIMENT

The experiment focused on EBL for the NAO robot, specifically examining a set of gestures associated with distinct emotional states, see Fig. 1 for the details. The primary objective was to design gestures that users could readily associate with specific emotions without the help of verbal communication.

### 4.1 EBL Design

In determining the most appropriate body gestures to represent emotions, we consulted the Github Emotional-gesturepapers collection [1] and then adjusted the gestures to align with those typically observed in Italian culture [24].

A total of eleven emotions were selected, with the first six being Ekman's basic emotions [12]: 1) Disgust; 2) Happiness; 3) Fear; 4) Anger; 5) Surprise; 6) Sadness.

Five additional well-known emotions were chosen from Plutchik's wheel [23] of emotions to to make the EBL more complete: 7) Love; 8) Interest; 9) Disapproval; 10) Boredom; 11) Thoughtfulness (Pensiveness).

As illustrated in Figure 1, the total number of proposed gestures exceeded eleven, as some emotions were represented by multiple gesture variants.

As introduced above, the primary sensory modality engaged during the experiment was vision, with sound intentionally excluded. This choice aimed to emphasize the role of visual perception in participants' interpretation of EBL conveyed through the robot's animations. Nonetheless, it is acknowledged that incorporating auditory cues might have facilitated faster emotion recognition.

To support emotion identification, a specific color was associated with each emotion and implemented through NAO's eye LEDs. The color selection process combined a rational approach, drawing on the symbolic meaning of colors in art, with more playful references, such as the animated film Inside Out [11], where Joy is associated with yellow, Sadness with blue, Anger with red, Fear with purple, and Disgust with green.

---

[1] https://github.com/mikecheninoulu/Emotional-gesture-papers

The gestures and behaviors required to express each emotion were programmed using NAO's development environment, specifically the NAOqi framework[1] and its Choregraphe multi-platform desktop application[2]. During the experiment, the animations were triggered manually by the experimenter using the same software.

### 4.2   Participants

The target group of the experiment described in this paper consisted of 20 participants aged between 71 and 89 years, 10 males and 10 females. Participants were recruited through collaboration with a care facility center and the University of Bari. The participants were self-sufficient elderly individuals residing in a care facility for older adults, with common age-related conditions such as hypertension, diabetes, hypertensive heart disease, and mobility impairments caused by spinal hernias. Informed consent was obtained from all participants prior to their involvement in the study. All data were collected anonymously and aggregated to ensure the protection of participants' privacy, in accordance with ethical guidelines for research with human subjects.

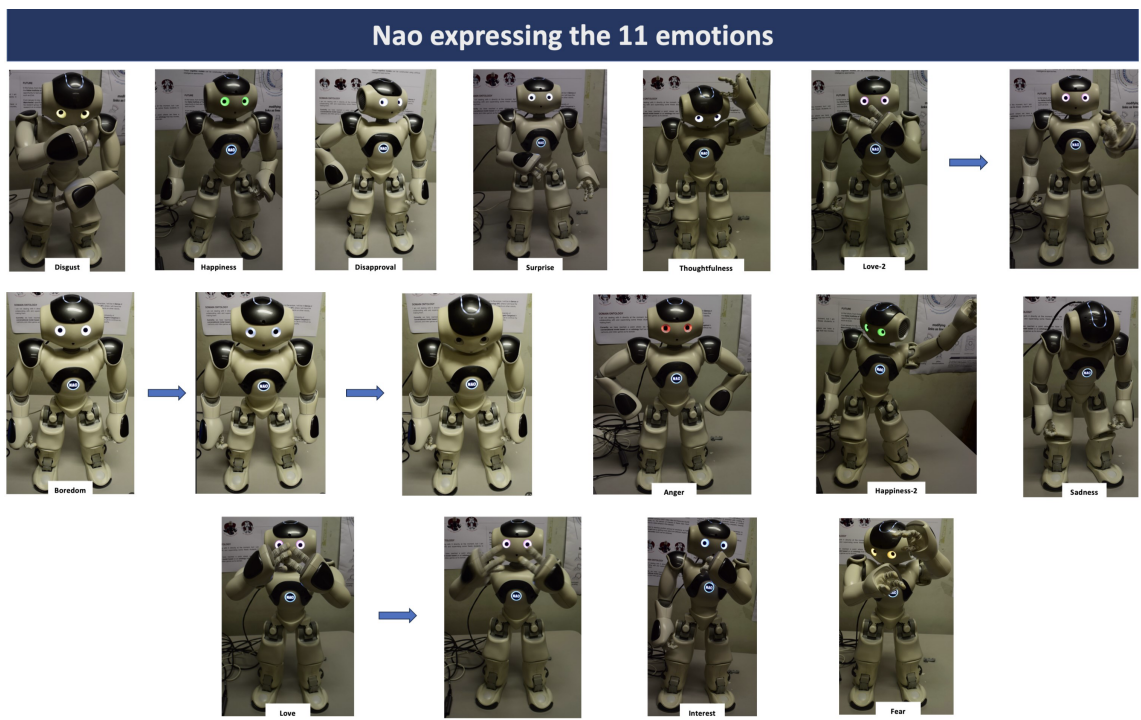

Fig. 1. Nao expresses the 11 emotions, some of them are performed twice.

---

[1] http://doc.aldebaran.com/1-14/dev/naoqi/index.html
[2] http://doc.aldebaran.com/1-14/software/choregraphe/choregraphe_overview.html

### 4.3  Procedure

We evaluated the ability of participants to recognize emotions conveyed through the NAO robot's body language. Each participant individually observed a sets of animations, each designed to represent one of eleven target emotions. Participants were provided with a predefined list of possible emotions to use as a reference. The animations were presented in a randomized sequence that differed from the order of the list. To minimize bias, participants were not informed that some animations were intentionally repeated.

After viewing each animation, participants were asked to identify the emotion they believed the robot was expressing by selecting from the reference list. Their responses were recorded for subsequent analysis. In cases of inconsistent responses, interventions were carried out by asking participants for suggestions or clarifications to improve interpretation.

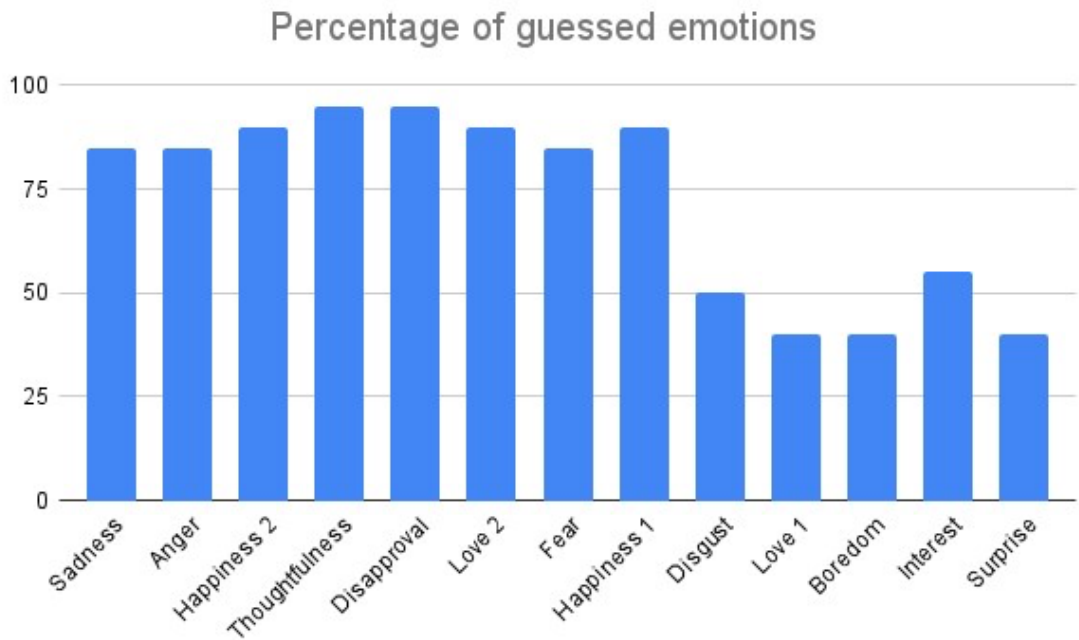

Fig. 2. Experimental results of the guessability study in the replicated experiment.

### 4.4  Results

Participants were able to recognize the emotions conveyed through the EBL of the NAO robot, as shown in Fig. 2. More in details the the obtained accuracy rates were the followings: Thoughtfulness (95%), Disapproval (95%), Happiness 2 (90%), Love 2 (90%), Happiness 1 (90%), Sadness (85%), Anger (85%), Fear (85%), Interest (55%), Disgust (50%), Love 1 (40%), Boredom (40%), and Surprise (40%). As it can be noticed, the following emotions were recognized in most cases, with recognition rates exceeding 80%: Sadness, Anger, Happiness 2, Thoughtfulness, Disapproval, Love 2, Fear, and Happiness 1. However, the following emotions were more difficult to recognize, as they were not identified correctly most of the time, with recognition rates ranging from 40% to 55%: Disgust, Love 1, Boredom, Interest, and Surprise, .

## 5 COMPARISON

Overall Group Differences in Emotion Recognition Accuracy

To evaluate whether overall emotion recognition accuracy differed across the three groups that performed the same experiments, namely young adults (students), children, and older adults, a Chi-square test of independence was conducted on the total number of correct versus incorrect responses across all 13 emotions.

The Chi-square test yielded a statistically significant result:

$$\chi^2(2, N = 2808) = 38.27, \quad p < .00000001$$

| Group | Correct | Incorrect | Total |
|---|---|---|---|
| Children (n=176) | 1569 | 719 | 2288 |
| Elderly (n=20) | 188 | 72 | 260 |
| Students (n=20) | 226 | 34 | 260 |

Table 1. Contingency table of correct/incorrect responses across groups (13 emotion items per participant).

| Group | Correct | Incorrect | Total |
|---|---|---|---|
| Children (n=176) | 68.6% | 31.4% | 100% |
| Elderly (n=20) | 73.3% | 27.7% | 100% |
| Students (n=20) | 86.9% | 13.1% | 100% |

Table 2. Contingency table reporting also percentage of correct/incorrect responses across groups (13 emotion items per participant).

This result indicates a significant association between group membership and overall accuracy in emotion recognition. Follow-up pairwise tests showed that:

- Students performed significantly better than both children and elderly (all $p < .001$).
- There was no significant difference between children and elderly participants ($p = .25$).

These findings suggest that students have significantly higher overall emotion recognition accuracy, whereas children and the elderly show comparable performance.

Post Hoc Pairwise Comparisons of Overall Accuracy

To examine which specific group differences accounted for the overall significant Chi-square result, we conducted post hoc pairwise Chi-square tests of independence between each group, along with effect size calculations using Cramér's V.

| Comparison | Chi-square | *p*-value | Cramér's V |
| --- | --- | --- | --- |
| Children vs. Students | 36.88 | < .000000001 | 0.12 |
| Children vs. Elderly | 1.35 | .25 | 0.02 |
| Elderly vs. Students | 16.22 | < .0001 | 0.18 |

Table 3. Post hoc pairwise Chi-square tests and effect sizes (Cramér's V) comparing overall accuracy across groups.

Looking at Table 3, we may observe the following:

- The comparison between students and children revealed a statistically significant difference with a small to moderate effect size;
- The comparison between students and elderly also yielded a significant result, with a moderate effect size;
- The difference between children and elderly was not statistically significant and showed only a negligible effect size.

These results confirm that students outperformed both children and elderly in overall emotion recognition accuracy, while the performance of children and elderly was statistically comparable.

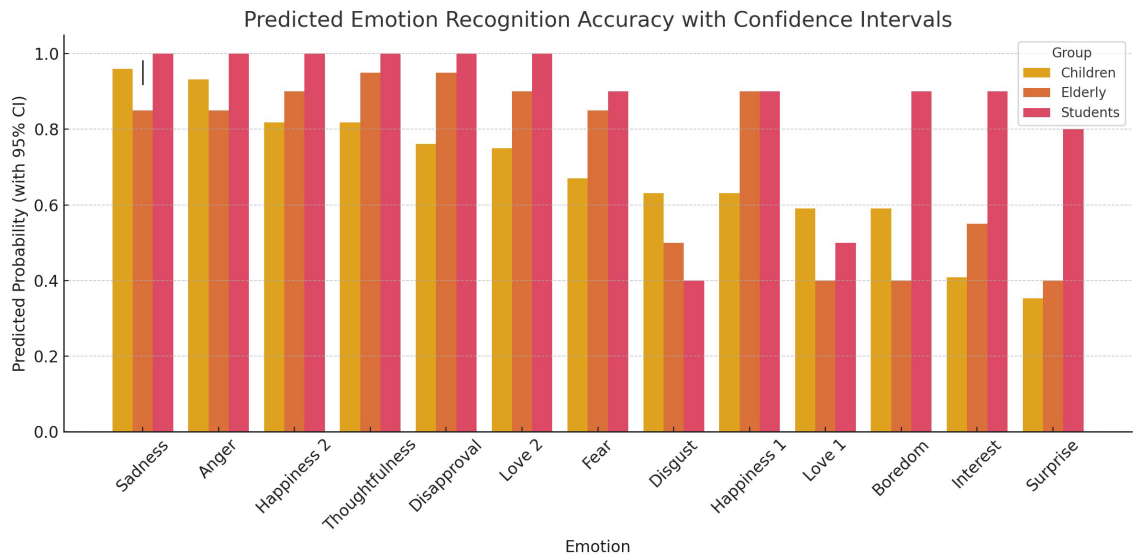

Fig. 3. The plot shows the predicted probabilities of correct emotion recognition. Includes error bars representing 95% confidence intervals for each Group × Emotion combination

### 5.1　　Predicted Probabilities of Emotion Recognition

Using a logistic regression model with interaction terms between group and emotion, we estimated the predicted probability of correctly recognizing each emotion for children, elderly participants, and students, see Fig. 5 for an overview. The analysis revealed systematic differences in recognition accuracy across both group and emotion:

- Students exhibited the highest predicted probabilities across nearly all emotions, with near-ceiling performance on basic emotions such as *Sadness*, *Anger*, and *Happiness*, and markedly superior recognition of complex emotions such as *Interest*, *Surprise*, and *Boredom*;
- Children showed lower predicted probabilities overall, particularly for more cognitively demanding emotions. The lowest probabilities were observed for *Surprise*, *Interest*, and *Love 1*, confirming these as the most difficult emotions to interpret for this group;
- Elderly participants generally performed better than children on some emotions (e.g., *Happiness 1*, *Disapproval*, *Thoughtfulness*), though their predicted probabilities were lower than those of students on several subtle emotions.

The model confirms an overall group effect, with students outperforming both children and elderly, and an emotion effect, with certain emotions being consistently more difficult to recognize regardless of group, as for instance Disgut and Love 1, suggesting a redesign for these gestures. The interaction terms highlight that the extent of group differences varies by emotion: for instance, performance gaps are especially pronounced for *Interest*, *Surprise*, and *Boredom*, but negligible for *Sadness* and *Anger*.

A bar plot with 95% confidence intervals, see Fig. 5, further illustrates these differences, highlighting the range of variability and the overlap (or lack thereof) in performance across groups and emotions.

5.2     Pairwise Comparisons of Emotion Recognition Accuracy Across Groups

To further investigate differences in emotion recognition accuracy across groups (children, elderly, and students), we conducted pairwise comparisons using z-tests for proportions, emotion by emotion. The results are summarized below, reporting only statistically significant differences ($p < .05$):

- Children vs. Students: Students performed significantly better than children on the following emotions: *Happiness 2* ($p = .037$), *Thoughtfulness* ($p = .037$), *Disapproval* ($p = .014$), *Love 2* ($p = .011$), *Fear* ($p = .035$), *Disgust* ($p = .045$), *Happiness 1* ($p = .016$), *Boredom* ($p = .007$), *Interest* ($p < .001$), and *Surprise* ($p < .001$).
- Elderly vs. Students: Students also outperformed elderly participants on: *Boredom* ($p = .001$), *Interest* ($p = .013$), and *Surprise* ($p = .010$).
- Children vs. Elderly: Elderly participants recognized *Sadness* ($p = .034$) and *Happiness 1* ($p = .016$) significantly better than children.

These results indicate that students consistently showed higher recognition accuracy, especially for more complex or subtle emotions, such as *Interest*, *Surprise*, and *Boredom*. Differences between children and elderly were fewer and limited to selected emotions.

CONCLUSION

This study investigated whether users across different age groups interpret emotional expressions differently when displayed by a humanoid robot (RQ2), and whether certain emotions are more easily recognized than others (RQ1).

Participants included children (n=176), elderly adults (n=20), and university students (n=20), each asked to identify 13 emotional expressions conveyed through the robot's body language.

Our findings confirm that emotion recognition varies significantly with age (RQ2). University students consistently demonstrated the highest accuracy across emotions, with performance significantly outperforming both children and elderly participants. Children and elderly showed comparable accuracy overall, although their confusion patterns varied by emotion. The lack of generalization across age groups suggests that EBL is interpreted in age-specific ways.

A further observation that merits attention is the fact that the principal designer of the gestures is a university student. Therefore, it can be hypothesized that her personal perspective on emotions may have had a bearing on their realization. Nevertheless, gestures were comprehended with remarkable ease by university students, who belong to the same age group as the designer. This may indicate that involving target users in the redesign of gestures could be beneficial in enhancing their recognition.

Concerning RQ1, Some emotions, particularly Sadness, Anger, and Happiness, were reliably recognized across all groups. In contrast, emotions such as Surprise, Interest, Boredom, and Love 1 were frequently misinterpreted, especially by children and elderly participants. These results partially replicate previous findings, though the recognition of some gestures (e.g., *Disgust*, *Love*) remained problematic across all samples.

These outcomes have clear implications for HRI. If robots are to express empathy and communicate effectively with diverse user populations, the emotional content of their gestures must be adapted to the stereotypical profiles of different age groups, similar the ones described in [15]. In particular:

- Designers should prioritize clear and unambiguous gestures for cognitively demanding emotions;
- Robots should adopt adaptive expressivity, modulating gesture intensity or clarity based on user feedback or demographics;
- Emotional communication should not assume uniform interpretation across users—customization is essential for inclusivity.

. In summary, this research underscores three distinct interpretative perspectives across age groups. The most accurate recognition was observed in young adults (students), while both children and elderly participants showed reduced performance and overlapping difficulties. These results support the hypothesis that age influences the interpretation of emotional body language in robots. Future studies should extend this investigation to include additional demographic segments—such as teenagers, middle-aged adults, or neurodiverse users—to more fully explore how age and individual differences shape the understanding of robot-expressed emotions.

ACKNOWLEDGMENTS

We would like to thank Matteo Giovanale for leading the educational robotics labs, and all the teachers and students who enthusiastically participated in the study.